\documentclass[letterpaper, 10 pt, conference]{ieeeconf}
\pdfoutput=1
\IEEEoverridecommandlockouts                              

\usepackage{graphicx}
\usepackage{mathtools}
\usepackage{url}
\usepackage{hyperref}
\usepackage{color}
\usepackage{multirow}
\usepackage{adjustbox}

\title{\LARGE \bf
Scoring Graspability based on Grasp Regression for Better Grasp Prediction
}

\author{Amaury Depierre$^{1,2}$, Emmanuel Dellandr\'ea$^{2}$ and Liming Chen$^{2}$
\thanks{$^{1}$Sil\'eane, Saint-Etienne, France
        {\tt\small a.depierre@sileane.com}}%
\thanks{$^{2}$University of Lyon, Ecole Centrale de Lyon, LIRIS, CNRS UMR 5205, France
        {\tt\small \{emmanuel.dellandrea, liming.chen\}@ec-lyon.fr}}%
}

\begin{document}

\maketitle
\thispagestyle{empty}
\pagestyle{empty}

\begin{abstract}

Grasping objects is one of the most important abilities that a robot needs to master in order to interact with its environment. Current state-of-the-art methods rely on deep neural networks trained to jointly predict a graspability score together with a regression of an offset with respect to grasp reference parameters. However, these two predictions are performed independently, which can lead to a decrease in the actual graspability score when applying the predicted offset. Therefore, in this paper, we extend a state-of-the-art neural network with a scorer that evaluates the graspability of a given position, and introduce a novel loss function which correlates regression of grasp parameters with graspability score. We show that this novel architecture improves performance from 82.13\% for a state-of-the-art grasp detection network to 85.74\% on Jacquard dataset. When the learned model is transferred onto a real robot, the proposed method correlating graspability and grasp regression achieves a 92.4\% rate compared to 88.1\% for the baseline trained without the correlation.

\end{abstract}

\section{INTRODUCTION}

Grasping objects is a crucial operation for many robot-aided applications: industry and logistics, household, human interaction, \textit{etc}. A reliable robotic grasp system could lead to a huge improvement in productivity, as well as new applications. Yet the performances achieved by current state-of-the-art systems are far from what a human can do. Humans can pick objects -known or unknown- of any shape in dark or bright lighting conditions with a success rate close to 100\%. To perform the same final action, a robotic system has to (1) analyze a sensor input to find a good grasp position, (2) plan a trajectory to reach this position, and (3) activate an end-effector to actually grasp the object. In this paper, we focus on the first part of the process and, more precisely, on detecting grasp positions for a parallel plate gripper in RGB images.

Early research works on grasp detection were based on analytic methods and 3D models \cite{miller2003automatic} \cite{bohg2010learning}. However, explicit object models are not always available in many real-world applications. The huge success of the paradigm of deep learning \cite{lecun2015deep}, especially in computer vision \cite{NIPS2012_4824} \cite{simonyan2014very}, inspires researchers to use convolutional neural networks (CNN) for the grasp prediction problem. In this case, the input of the network is a sensor image, either RGB, RGB-D, RGD (RGB image where the blue channel has been replaced by depth information) or depth only. Despite being more noisy than 3D models, these data are more readily available and thereby allow effective training of deep neural networks (DNN). Once trained, these DNN models  enable real-time processing \cite{redmon2015real}, while ensuring good grasp performance \cite{lenz2015deep}.

\begin{figure}
	\centering
	\def\svgwidth{0.4\textwidth}
	\Large
	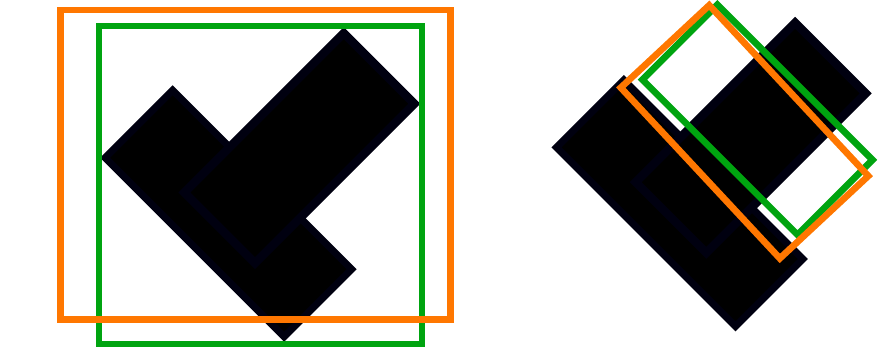
	\caption{(a) Two axis-aligned object detection boxes (b) two oriented grasp detection boxes. The loss between the green and the orange boxes is the same in (a) and (b). In both cases, the visual input based score is the same, however in (b) the green grasp is good but the orange is not, due to a collision with another part of the object.}
	\label{Fig1}
\end{figure}

The current literature features two different kinds of approaches for grasp prediction from images using DNN: generative methods and evaluation methods. With generative methods, a DNN is trained to predict one (or multiple ordered) grasps from the sensor input. Predictions can be the direct regression of  grasp parameters  \cite{redmon2015real} or deformation values \textit{w.r.t.} a reference grasp \cite{guo2017hybrid} \cite{zhou2018fully} \cite{chu2018real}, also called an anchor. In evaluation methods, grasp proposals are presented to the network, whose role is to order them in terms of grasp quality \cite{mahler2017dex} \cite{mahler2017suction}. While evaluation methods are classification only and therefore generally simpler than generative methods, they cannot perform end-to-end grasp prediction and have to rely on another method to propose the grasps they evaluate. 
On the other hand, generative methods predict a score from an input scene image and simultaneously, but independently, regress the offset of reference anchor boxes. This approach works well in practice for object detection and has been adapted to grasp detection leading to state-of-the-art performances. For object detection, even if the predicted box does not exactly match the ground truth, the visual input contains all the information needed to determine whether or not the object is present. However, this is not always true for grasp detection with an oriented box: an error in the prediction can have much more impact for the outcome of the grasp, as can be seen in \hyperref[Fig1]{Fig. 1 (b)}. Visual information is not always sufficient to determine the grasp result, and the grasp parameters could be required to enable good prediction of the graspability of a given grasp. To overcome this issue, we propose to combine the best of the aforementioned generative and evaluation approaches, and introduce a novel method that not only performs the regression of grasp parameters of the object in an input image, but also explicitly evaluates the graspability of a given predicted grasp in extending a state-of-the-art grasp prediction network with a scorer network optimized through a novel loss function in particular.

In this paper, we present two contributions:

\begin{enumerate}
    \item We  propose a novel DNN architecture, which uses its grasp quality evaluation to improve grasp regression through a newly introduced loss;
    \item We perform extensive experiments and show that the proposed grasp detection method significantly outperforms the state-of-the-art both on Jacquard \cite{depierre2018jacquard} and on a real-life robotic application;
\end{enumerate}

The rest of this paper is organized as follows. \hyperref[secII]{Section II} overviews the related work. \hyperref[secIII]{Section III} explains the representation we used for robotic grasps. \hyperref[secIV]{Section IV} presents in detail our network architecture, while \hyperref[secV]{Section V} discusses the experiments we conducted to evaluate our architecture. \hyperref[secVI]{Section VI} concludes this paper.

\section{RELATED WORK}
\label{secII}

Past works \cite{miller2003automatic} \cite{bohg2010learning} on grasp detection used 3D models and analytical methods to analyze scenes and find grasp candidates. In real case applications with unknown objects or environments, perfect knowledge of these models is difficult to acquire. For that reason, research was later oriented towards image analysis. More specifically, Lenz \textit{et al.} \cite{lenz2015deep} used a sliding window approach and a deep neural network to classify image patches and extract potential grasp locations. They reached an accuracy of 73.9\% on the Cornell Grasping Dataset \footnote{\url{http://pr.cs.cornell.edu/grasping/rect_data/data.php}}. In \cite{redmon2015real}, Redmon \textit{et al.} improved both accuracy and processing time by using direct regression. Their network predicts $5$ parameters describing the grasp, as well as a positive outcome probability for each spatial area in a $N\times N$ grid covering the image. This method raised accuracy to 88.0\%. Using deeper neural networks like ResNet \cite{He2015} only marginally increased the performances in grasp detection to 89.21\% in \cite{kumra2017robotic}. Networks with orders of magnitude less weights \cite{morrison2019learning} were also developed, achieving similar accuracy values.

Reinforcement learning methods have also been developed to train robots to move towards a good grasp location. This kind of approach requires a much larger volume of data that can be either collected with real robots over several days \cite{pinto2016supersizing} or in simulation, requiring transfer learning to adapt the trained model on real images \cite{james2019sim}.

Inspired by research in object detection \cite{ren2015faster} \cite{liu2016ssd} \cite{redmon2016you}, Guo \textit{et al.} \cite{guo2017hybrid} and Chu \textit{et al.} \cite{chu2018real} introduced the notion of reference anchor box. With anchor boxes, the neural network is not trained to directly regress a grasp but instead to predict a deformation of a reference box. This simplifies the regression problem by introducing prior knowledge on the size of the expected grasps. In \cite{guo2017hybrid}, the reference box is axis aligned, and the network also produces a quality score and an orientation as classification between discrete angle values. In \cite{chu2018real}, the quality score is included in the orientation classification, letting the network predict both grasp regression values and the discrete orientation classification score. This network achieved an accuracy of 96.1\% on the Cornell Grasp Dataset.

Zhou \textit{et al.} proposed in \cite{zhou2018fully} to remove the orientation classification by introducing oriented anchor boxes. Instead of having multiple scales or aspect ratios for the reference grasps, the authors used only one anchor box with multiple orientations. Then, the angle of the grasp is predicted  with respect to this reference orientation, just like the position and size values. Thus, the network predicts five regression values and one grasp quality score for each oriented reference box at each location in the feature map and achieves a state-of-the-art accuracy of 97.74\%. However, grasp quality prediction only depends on the image information and is not directly correlated to the regression prediction. Our method extends this approach by adding a direct dependency between the regression prediction and the score evaluation.

\section{PROBLEM STATEMENT}
\label{secIII}

\begin{figure}
	\centering
	\def\svgwidth{0.4\textwidth}
	\LARGE
	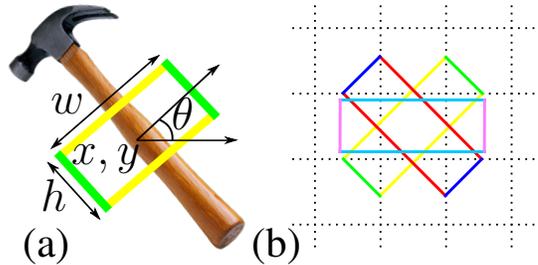
	\caption{(a) Example of a 5D representation of a grasp (b) Three examples of oriented anchor boxes with the same dimensions and angles of -45$^{\circ}$, 0$^{\circ}$ and 45$^{\circ}$ centered on the same pixel of the feature map}
	\label{Fig2}
\end{figure}

In this paper, we consider the problem of detecting successful grasp positions from RGB images of various objects lying on a plane. As the network does not have any space information, we use a 2D representation for the grasp rather than a 3D one. As shown in \hyperref[Fig2]{Fig. 2 (a)}, each grasp is represented as:
\begin{equation}
    \label{eq1}
    g=\{x,y,h,w,\theta\}
\end{equation}
where $(x,y)$ are the position of the center, $(h,w)$ its dimensions, and $\theta$ its orientation. In \cite{lenz2015deep} Lenz \textit{et al.} showed that this representation works well in practice, even with real robotic systems.

To simplify the regression problem, prior knowledge about the position, size and orientation of the grasp is introduced through oriented reference grasps \cite{zhou2018fully}. These reference grasps, also called anchor boxes, are defined as $g_a = (g_{ax}, g_{ay}, g_{aw}, g_{ah}, g_{a\theta})$. The grasp is then defined as a deformation $\delta = (\delta_x, \delta_y, \delta_w, \delta_h, \delta_\theta)$ of a reference grasp according to the following equation:

\begin{equation}
    \label{eq2}
    \begin{array}{rcl}
        x & = & \delta_x * g_{aw} + g_{ax}  \\
        y & = & \delta_y * g_{ah} + g_{ay}  \\
        w & = & \exp(\delta_w) * g_{aw} \\
        h & = & \exp(\delta_h) * g_{ah}  \\
        \theta & = & \delta_\theta * (180 / k) + g_{a\theta}
    \end{array}
\end{equation}
\noindent where k is the number of different anchor boxes. \hyperref[Fig2]{Fig. 2 (b)} illustrates three different examples of oriented anchor boxes.

\section{PROPOSED METHOD}
\label{secIV}
We first describe the proposed network architecture (sect. \ref{subsecNetworkArchitecture}), then introduce the loss function (sect.\ref{secLossFunction}) and finally present the training procedure (sect.\ref{subsectionTrainingProcedure}).

\subsection{Network Architecture}
\label{subsecNetworkArchitecture}

\begin{figure*}
	\makebox[\textwidth][c]
	{
	    \def\svgwidth{1.0\textwidth}
	    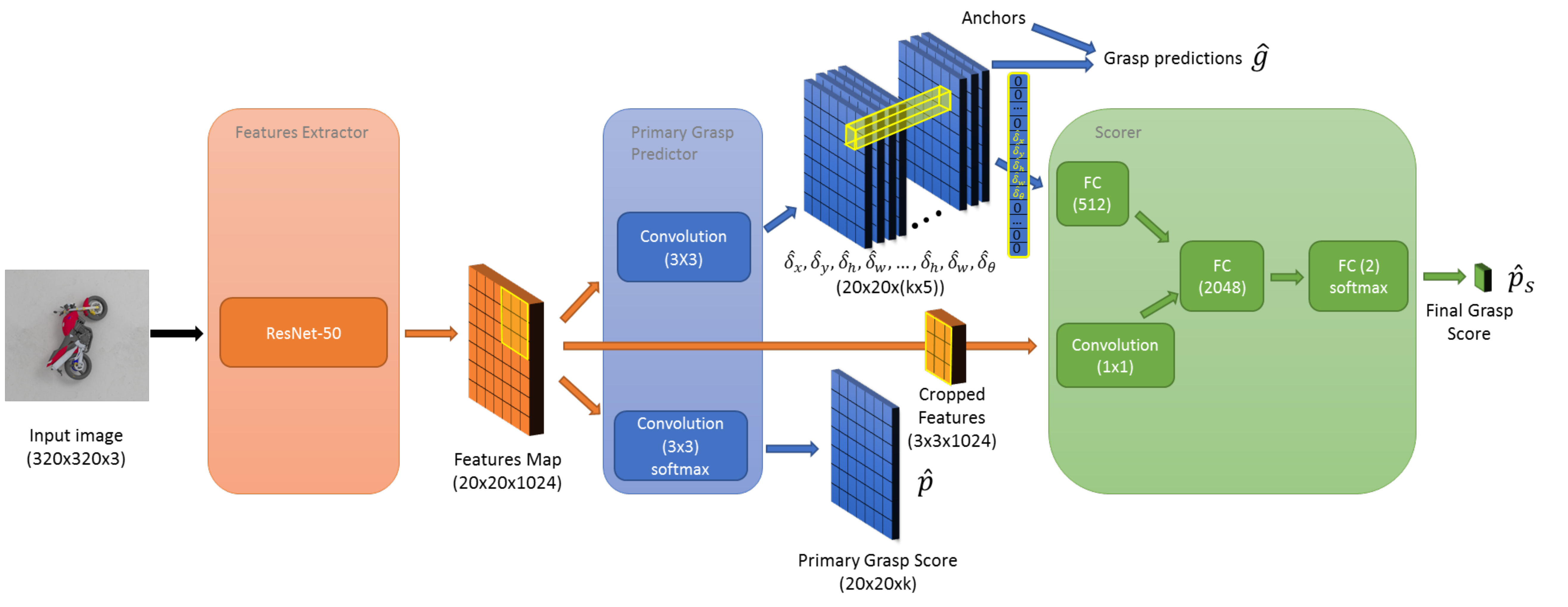
	}
	\caption{Global view of our architecture and its three components, the feature extractor, the intermediate grasp predictor, and the scorer network}
	\label{Fig3}
\end{figure*}

\hyperref[Fig3]{Fig. 3} presents the global architecture of our network. It contains three main components: a feature extractor (FE), a primary grasp predictor (PGP) and our newly introduced scorer. Given a feature delivered by FE from an input image, PGP aims to predict grasp parameters along with a primary grasp score indicating the quality of the corresponding grasp. This primary grasp score is predicted independently from the grasp parameters, although the same feature map is used as input. As a result, the scorer refines this primary grasp score and delivers a final one for each set of predicted grasp parameters and the visual neighborhood of the corresponding grasp position.   

\paragraph{Feature Extractor (FE)}

For all our experiments, we used the popular ResNet-50 network \cite{He2015} as a feature extractor. As we are using the ResNet as a fully convolutional encoder, input image size is not restricted to the pre-trained $224 \times 224$ size. Depending on the desired size of the output feature map, any convolution result in the ResNet can be used as an output feature map. In our experiments, we used an image input size of $320\times 320$ pixels and took the output of the fourth convolution block. Therefore, the output feature maps have a size of $20 \times 20\times 1024$.

\paragraph{Primary Grasp Predictor (PGP)}

The second part of our network is an oriented anchor box-based grasp detector. We used the state-of-the-art architecture presented by Zhou \textit{et al.} \cite{zhou2018fully} with two heads. Specifically, two separate convolutional layers are added to predict separately from an input feature map, for each pixel of the feature map and for each reference anchor box, five regression values $\widehat{\delta} = (\widehat{\delta}_x, \widehat{\delta}_y, \widehat{\delta}_w, \widehat{\delta}_h, \widehat{\delta}_\theta)$, as well as a primary grasp quality score $\widehat{p}$. This primary grasp quality score shows the quality of the corresponding oriented reference grasp: a score close to 1 means a high confidence of a good location, while a score close to 0 indicates an inadequate position or orientation for a parallel plate gripper.  It is important to note that this primary grasp prediction network does not have the anchors as an input: it only predicts a deformation that is afterwards applied to the corresponding anchor.

\paragraph{Scorer}

The primary grasp quality score, predicted by the previous PGP, only depends on the visual features and not on the final predicted grasp parameters $\widehat{g}$. So a reference grasp on a good location could have a high score despite being a bad prediction, once the final grasp $\widehat{g}$ is computed through \hyperref[eq2]{Eq. 2}. Moreover, the score estimation cannot be used to improve regression quality. To deal with these issues, we extend this state-of-the-art network by a third component: a scorer network. Inspired by two stage object detection architectures \cite{ren2015faster} this scorer is implemented on the top of PGP. However, unlike in object detection, its role is not to refine the predicted grasp parameter by PGP but its grasp quality score. Similarly to the Grasp Quality CNN used in different versions of Dex-Net \cite{mahler2017dex} \cite{mahler2017suction} \cite{satish2019policy}, the scorer network predicts the likelihood $\widehat{p}_s$ that a proposed grasp is a good one, using not only visual information but also the actual predicted grasp parameters. 

In \hyperref[Fig3]{Fig. 3}, we can see the detailed implementation of this scorer network. We kept this network small to avoid adding too many computation costs and memory usages. It takes as input the visual neighborhood of a grasp position along with a set of estimated grasp parameters. Specifically, a $3\times 3$ area from the feature map around the grasp position is sent through a $1\times 1$ convolutional layer with $1024$ filters. Its output is then flattened to a vector of size 9216 ($3\times 3\times 1024$). As for each potential grasp position, there exist $k$ anchors with the corresponding grasp parameters, we encode the input of grasp parameters for a given grasp  as a vector of size $5\times k$, where every value is set to $0$ except the $5$ values corresponding to the considered anchor box, which are set to the $\widehat{\delta}$ output from PGP. Keeping a $5\times k$ dimension vector allows the network to differentiate the k base anchors, while not having to transform the $\widehat{\delta}$ coordinates with \hyperref[eq2]{Eq. 2} between the PGP and the Scorer. This $5\times k$ vector is passed through a $512$ neuron fully connected layer, and the result is concatenated with the image vector. The result is a $9728$ dimensional vector processed by two last fully connected layers, resulting in a graspability score and a non-graspability score. These two scores are then processed by a softmax to get a grasp success likelihood. All the layers of the scorer network (except the last one) are followed by a leaky ReLU with a negative slope set to $0.1$.

In short, our whole model with its three components has three outputs: a grasp parameter prediction from PGP, a primary grasp quality score from PGP describing the likelihood that the corresponding reference anchor box is a good grasp, and a final score from Scorer delivering the likelihood that the actual predicted grasp is a good one.

\subsection{Loss functions}
\label{secLossFunction}

There are three different loss functions in our architecture. Loss functions for the prediction of the primary grasp quality score by PGP and the refined grasp quality score from the scorer are both softmax cross-entropy losses. They are both computed only on a subset of the predictions: $P$ positive and $3P$ negative using PGP, and $T$ grasps for the Scorer.
\begin{equation}
    \label{eq3}
    \begin{split}
        \mathcal{L}_{intermediate}(\widehat{p}) = - \frac{1}{4P} \sum_{i = 1}^{4P} p_i\log(\widehat{p}_i) \; + \\ (1-p_i))\log(1-\widehat{p}_i)
    \end{split}
\end{equation}

\begin{equation}
    \begin{split}
        \mathcal{L}_{scorer}(\widehat{\delta}, \widehat{p_s}) = - \frac{1}{T} \sum_{i = 1}^{T} p_{si}(\widehat{\delta}_i)\log(\widehat{p}_{si}) \; + \\ 
        (1 - p_{si}(\widehat{\delta}_i))\log(1 - \widehat{p}_{si})
    \end{split}
\end{equation}

\noindent where $p_i$ (resp. $p_{si}(\widehat{\delta}_i)$) is the ground truth associated with the $i^{th}$ primary score prediction (resp. scorer score prediction) calculated as explained in \ref{subsubsectionAnchorSelection}.

The regression loss for the prediction of grasp parameters is composed of two terms: a classic smooth L1 function to ensure the predicted grasps match the annotated ground truth, and a newly introduced second term using the scorer output to guide the gradients in a direction that improves the estimated quality of the predicted grasp.
\begin{align}
    \begin{split}
        \mathcal{L}_{reg}(\widehat{\delta}, \widehat{p}_s) = {} &\frac{\alpha}{P} \sum_{i = 1}^{P} \sum_{m\in\{x,y,w,h,\theta\}} L1_{smooth}(\delta_{mi} - \widehat{\delta}_{mi}) \; \\ &- \frac{1}{T}\sum_{i = 1}^{T} \log(\widehat{p}_{si})
    \end{split}
\end{align}

\noindent where $\delta_i$ are the ground truth values obtained from the annotated grasp inverting \hyperref[eq2]{Eq. 2}. To be consistent with \cite{zhou2018fully} we kept $\alpha=2$ in our training.

As we do not want to update PGP's weights to predict grasps that are easier to classify for the scorer (i.e. reducing $\mathcal{L}_{scorer}$), gradients from $\mathcal{L}_{scorer}$ are only used to update the scorer and the FE networks. Similarly, gradients from $\mathcal{L}_{reg}$ are not used to update the scorer network but only PGP and the FE networks.

\subsection{Training Procedure}
\label{subsectionTrainingProcedure}

\subsubsection{Anchor selection}
\label{subsubsectionAnchorSelection}

As all the grasp parameter predictions are generated with respect to reference anchors, choosing them correctly is crucial for the performance of the whole network. In \cite{chu2018real}, Chu \textit{et al.} used $3$ different scales and aspect ratios for a total of $9$ axis-aligned anchors. However, Zhou \textit{et al.} showed in \cite{zhou2018fully} that orientation is more important for accuracy than the dimensions of the box. Therefore, we also used for our experiments only one anchor with $k = 6$ different orientations. 

Instead of using a ratio of 1:1 with an arbitrary size, before the training process we compute a mean box of all the ground truth grasps from the training dataset. This approach gives us values for $h$ and $w$ of the anchor boxes and has been proven to yield good results in object detection \cite{redmon2017yolo9000}. As grasps usually have a larger $w$ than $h$, using a mean grasp helps as it provides network reference grasps closer to the real ones than when using a 1:1 ratio, while also reducing the number of hyperparameters needing manual optimization.

\subsubsection{Grasp selection}
\label{subsubsectionGraspSelection}
The primary score of PGP is trained using the fast but less accurate Angle Matching strategy presented in \cite{zhou2018fully}: the ground truth score $p$ is set to $1$ if the distance between the angles of the corresponding anchor and an annotated ground truth grasp is under a threshold and set to $0$ otherwise. The scorer network uses Jaccard Matching instead: the ground truth score $p_s$ of a predicted deformation $\widehat{\delta}$ is set to $1$ if the corresponding grasp $\widehat{g}$ calculated through \hyperref[eq2]{Eq. 2} is both close in orientation and has an intersection over union with a ground truth grasp over a threshold. As this is more time-consuming than Angle Matching, not all the $20\times 20\times k$ predictions are evaluated during training. Only a subset of all the predictions is used to compute the gradients involving our scorer network. To determine which of the grasps are selected, we use the primary score delivered by PGP only based on the visual input. The $T$ anchor boxes with the highest scores are selected and only the corresponding $T$ grasps are evaluated by the scorer network and used for training. At test time, all the grasps are evaluated by the scorer.

\section{EXPERIMENTS AND RESULTS}
\label{secV}
We first define the experimental setup (sect.\ref{subsectExperimentalSetup}), then analyze the experimental results (sect.\ref{subsectResults}), and finally present the testing results on a real robot (sect.\ref{subsectRealRobotTesting}).

\subsection{Experimental setup}
\label{subsectExperimentalSetup}

\subsubsection{Datasets}
In order to evaluate the performance of our model, we used the widely adopted Cornell Grasping Dataset. State-of-the-art models achieve very high accuracy rates on this dataset, and the few unsuccessful detection results are in fact visually consistent with good grasp locations, even if they do not match any ground truth grasp with Jaccard Matching. For this reason, we also used the larger Jaquard Dataset \cite{depierre2018jacquard}, which has 54k images in comparison with the 1k of the Cornell Dataset. Just as in the previous work, we divided the dataset into 5 parts and performed cross-validation. This separation was carried out object-wise, which means that images containing one object are either all in the training dataset or all in the testing one. Presented performances are the averaged accuracy over the five tests.

\subsubsection{Training and evaluation details}
As overfitting is a common issue with neural networks, especially when fully connected layers are deployed, we used online data augmentation to make sure the network does not see twice the exact same image during the whole training process. In detail, the RGB image is randomly rotated around its center, mirrored, shifted up to $50$ pixels in both axes, and finally rescaled to $320\times 320$ pixels before feeding to the network. This augmentation is not performed at testing time.

To help the training, all the weights of the feature extractor are initialized from a pretraining phase on the large RGB dataset ImageNet \cite{deng2009imagenet}. Weights for the other layers are randomly initialized. The network is trained through Stochastic Gradient Descent with a momentum of $0.9$ for 100k iterations. Batch size is set to $10$ for all the models. The learning rate is set to $0.001$ and the weight decay to $0.0001$. To train the scorer network, $T$ is set to $64$ as we found this was a good trade-off to balance positive and negative examples, as well as being a value small enough to avoid overly increasing training time. To train the scorer, we used Jaccard Matching with thresholds of 15$^{\circ}$ and 25\% for the intersection over union. For evaluation, we used the more common 30$^{\circ}$ criterion (and 25\% for the intersection over union) to be consistent with previous work.

\subsection{Results}
\label{subsectResults}
Once trained with the whole architecture (FE, PGP and Scorer), our proposed model delivers from an input RGB image 3 outputs: regressed grasp parameters for each potential grasp position by PGP, a primary grasp quality score from PGP, and a refined final one from Scorer for each anchor on that position.  
To evaluate the influence of our scorer network on performances, we compared the success rate of multiple architectures on both the Cornell and Jacquard datasets. For our model, we evaluated accuracy using both the primary grasp probability from PGP and the final refined one from the Scorer, although they were trained together during the training procedure. 
The results are presented in \hyperref[tableI]{Table I}. To obtain a baseline of comparison with a similar architecture to ours without the scorer network, we also trained the algorithm proposed by Zhou et \emph{al}. in \cite{zhou2018fully} on the Jacquard Dataset.

{
    \renewcommand{\arraystretch}{1.4}
	\begin{table}[]
	    \caption{Grasp Detection Accuracy}
		\label{tableI}
	    \begin{adjustbox}{width=\columnwidth,center}
    		\begin{tabular}{|c|c|c|c|}
    		\hline
    		\multirow{2}{*}{\textbf{Algorithm}} & \multirow{2}{*}{\textbf{Backbone}} & \multicolumn{2}{c|}{\textbf{Dataset}} \\ \cline{3-4} 
     &  & \textbf{Cornell} & \textbf{Jacquard} \\ \hline
    Depierre et \emph{al}. \cite{depierre2018jacquard} & AlexNet & 86.88\% & 74.21\% \\ \hline
    Chu et \emph{al}. \cite{chu2018real} & ResNet-50 & 95.5\% & - \\ \hline
    \multirow{2}{*}{Zhou et \emph{al}. \cite{zhou2018fully} } & ResNet-50 & 94.91\% & 82.13\% $\pm$ 0.36* \\ \cline{2-4} 
     & ResNet-101 & 96.61\% & - \\ \hline
     \multirow{2}{*}{Morrisson et \emph{al}. \cite{morrison2019learning} } & GG-CNN & 88\% & 78\% \\ \cline{2-4} 
     & GG-CNN2 & - & 84\% \\ \hline
    Ours (using primary score) & ResNet-50 & 95.02\% $\pm$ 0.32 & 83.61\% $\pm$ 0.41 \\ \hline
    Ours (scorer) & ResNet-50 & 95.2\% $\pm$ 0.19 & 85.74\% $\pm$ 0.17\\ \hline
    \multicolumn{4}{l}{*our implementation}
    		\end{tabular}
		\end{adjustbox}
	\end{table}
}

As the results show, our proposed architecture performs similarly to the state-of-the-art on the Cornell Dataset with 95.2\% compared to 95.5\% for the network from \cite{chu2018real} with the same ResNet-50 backbone, and is only 1.4 points behind the performance of \cite{zhou2018fully} with a much larger ResNet-101 backbone. On the Jacquard dataset, our architecture achieves state-of-the-art performances with an accuracy of 85.74\%, outperforming the previous state-of-the-art performance by 1.74 points. Compared to a baseline at third row with a similar architecture (FE, PGP) but without the scorer part, our model performs 3.61 points better, showing that the scorer and its associated loss are useful for the grasp prediction problem.
By comparing the last two lines of \hyperref[tableI]{Table I}, we also observe that the grasp quality score predicted by the scorer, based on the visual input and the grasp parameter prediction, is more accurate than the primary one, based only on the visual information.

\subsubsection{Anchor size influence}

In \cite{zhou2018fully}, the authors used only one anchor of size $54$x$54$ with multiple orientations. To evaluate the influence of this parameter on the final accuracy of our model, we trained it on the Jacquard dataset with different anchor sizes and ratios. The results are presented in \hyperref[tableII]{Table II}. We can deduce from these values that anchor size is not the most crucial hyperparameter for this kind of approach, with an accuracy of 84.06\% for the worst network and 85.74\% for the best one with the mean anchors. Using the mean box does not only provide the best performance, but also the fastest training as it does not need an intensive grid-search to optimize this hyperparameter.

\begin{table}[]
\caption{Accuracy of our model for different anchor sizes on the Jacquard Dataset}
\label{tableII}
    \begin{adjustbox}{width=\columnwidth,center}
        \begin{tabular}{|c|c|c|c|c|c|}
        \hline
        \textbf{Anchor size} & 54x54 & 54x27 & 108x54 & 108x27 & Mean (91x26) \\ \hline
        \textbf{Accuracy} & 84.62\% & 85.39\% & 84.06\% & 84.53\% & 85.74\% \\ \hline
        \end{tabular}
    \end{adjustbox}
\end{table}

\subsubsection{Grasp selection influence}

To evaluate the effect of the selection of the $T$ grasp parameter predictions using the primary score, we compared our full model to the same architecture but trained without this selection process: at training time, all the $20 \times 20 \times k$ grasp proposals were processed by the scorer during the forward pass, and the $T$ with the highest final score were used for back-propagation (instead of the $T$ with the highest primary score). This model, in addition to being slower and consuming more memory during training, only has an accuracy of 82.36\%, 3.38 points less than the model using primary score predicted by PGP. This shows that using the primary quality score to train the scorer on adversarial examples (which are potentially good based on the visual input but not necessarily good once the real criterion is evaluated) helps the network to converge towards a better solution.

\subsection{Real Robot Testing}
\label{subsectRealRobotTesting}

\begin{figure}
    \centering
    \def\svgwidth{0.3\textwidth}
    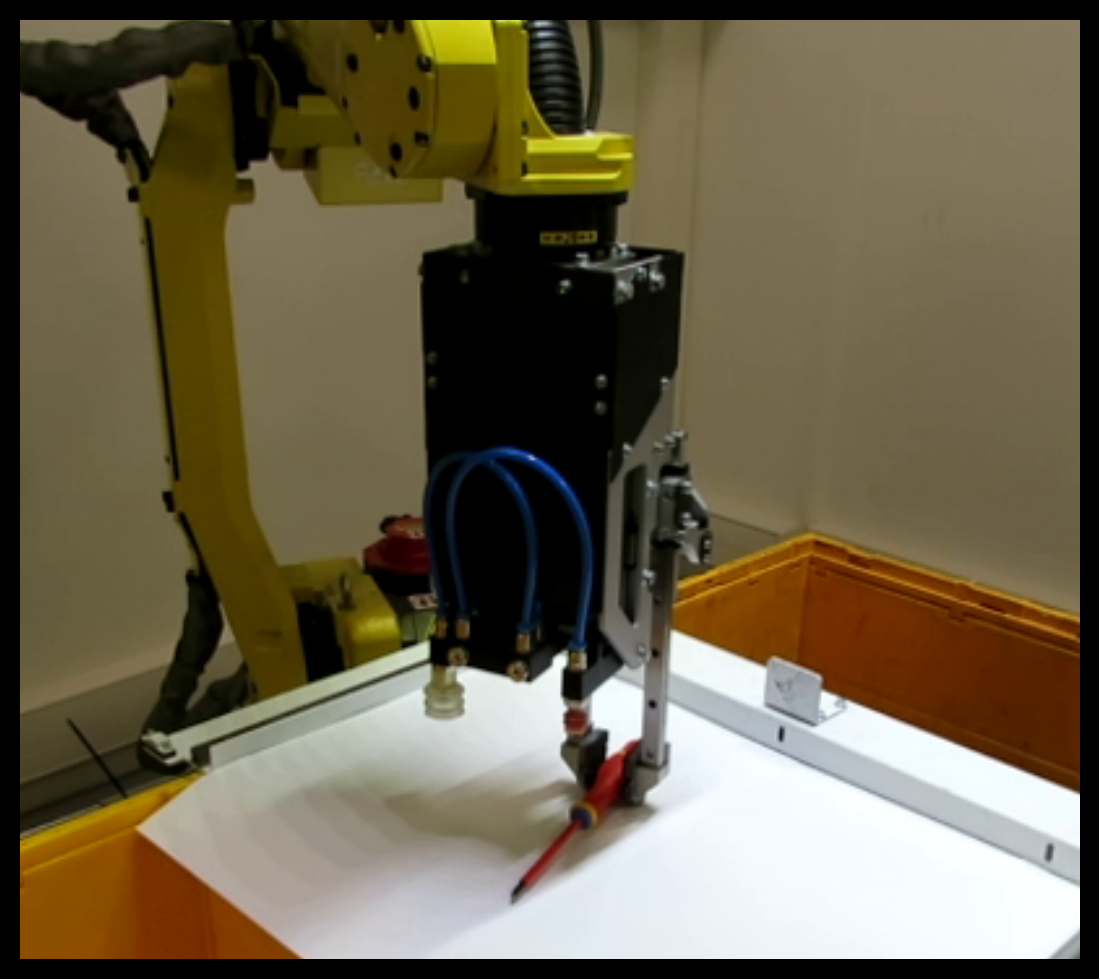
	\caption{Our robot performing a grasp with its parallel plate gripper. The top view RGB camera is located above the gripping area}
	\label{Fig4}
\end{figure}

One important question when training models on datasets is how they can generalize on real applications. Therefore, to evaluate this reality gap, we tested our model on a Fanuc's robotic arm with a parallel plate gripper and a set of various objects. We used 15 common household objects and 10 industrial parts to ensure a wide diversity in shapes, colors and materials. \hyperref[Fig2]{Fig. 4} shows our physical setup performing a grasp on one of the household objects. We considered a grasp was successful when the robot could lift the object, move it away and put it back on the gripping area without dropping it in the middle.
With this setup, we achieved an accuracy of 88.1\% for our baseline without the scorer and 92.4\% for our full model. Both were trained on the full Jacquard Dataset. This 4.3 point difference shows that not only our  architecture performs better on the dataset it was trained on but also that it has a better internal grasp representation, allowing it to be more accurate in real case applications.

\section{CONCLUSIONS}
\label{secVI}

In this work, we presented a new architecture combining grasp regression and score evaluation, which uses grasp quality score estimation to improve regression quality. We evaluated it on both Cornell and Jacquard datasets, as well as on a real robotic setup. The experimental results show that our model with the added scorer is more efficient for predicting grasps than previous state-of-the-art architectures. Our future work will focus on new methods like meta-learning to allow grasp prediction models to quickly adapt to new situations.

\addtolength{\textheight}{-21cm}

\section*{Acknowledgment}

This work was supported in part by the EU FEDER funding through the FUI PIKAFLEX project, by the French National Research Agency, l'Agence Nationale de Recherche (ANR) through the ARES LabCom under grant ANR 16-LCV2-0012-01, as well as the LEARN-REAL project within the the EU CHIST-ERA program under grant ANR-18-CHR3-0002-01 and Chiron project within the trilateral France-Germany-Japan program on AI under grant ANR-20-IADJ-0001-01 for Liming Chen and Emmanuel Dellandréa.






\bibliographystyle{IEEEtran}
\bibliography{IEEEabrv,biblio}

\end{document}